\documentclass[10pt,twocolumn,letterpaper]{article}

\usepackage{iccv}

\usepackage{times}
\usepackage{epsfig}
\usepackage{graphicx}
\usepackage{amsmath}
\usepackage{amssymb}
\usepackage{biblatex}
\usepackage{listings}
\usepackage{algorithm}
\usepackage{stfloats}
\usepackage{algorithm}
\usepackage[noend]{algpseudocode}
\usepackage{subcaption}
\usepackage{amsthm}
\usepackage{tabu}
\usepackage[usenames, dvipsnames]{color}
\usepackage{csvsimple}
\usepackage[autolanguage]{numprint}

\newcommand{\e}{\mathrm{E}}
\newcommand{\var}{\mathrm{Var}}

\newtheorem{theorem}{Theorem}
\usepackage{pdfpages}

\usepackage[breaklinks=true,bookmarks=false]{hyperref}

\iccvfinalcopy % *** Uncomment this line for the final submission

\def\iccvPaperID{****} % *** Enter the ICCV Paper ID here

\begin{document}
    
\title{Deep Residual Learning in the JPEG Transform Domain}

\author{Max Ehrlich and Larry Davis\\
    {\tt\small maxehr@umiacs.umd.edu} \qquad {\tt\small lsd@umiacs.umd.edu}\\
    University of Maryland, College Park, MD, USA.
}

\maketitle

\begin{abstract}   
    We introduce a general method of performing Residual Network inference and learning in the JPEG transform domain that allows the network to consume compressed images as input. Our formulation leverages the linearity of the JPEG transform to redefine convolution and batch normalization with a tune-able  numerical approximation for ReLu. The result is mathematically equivalent to the spatial domain network up to the ReLu approximation accuracy. A formulation for image classification and a model conversion algorithm for spatial domain networks are given as examples of the method. We show skipping the costly decompression step allows for faster processing of images with little to no penalty in the network accuracy.
\end{abstract}

\section{Introduction}

The popularization of deep learning since the 2012 AlexNet \cite{krizhevsky2012imagenet} architecture has led to unprecedented gains for the field. Many applications that were once academic are now seeing widespread use of machine learning with success. Although the performance of deep neural networks far exceeds classical methods, there are still some major problems with the algorithms from a computational standpoint. Deep networks require massive amounts of data to learn effectively, especially for complex problems \cite{najafabadi2015deep}. Further, the computational and memory demands of deep networks mean that for many large problems, only large institutions with GPU clusters can afford to train from scratch, leaving the average scientist to fine tune pre-trained weights.

This problem has been addressed many times in the literature. Batch normalization \cite{ioffe2015batch} is ubiquitous in modern networks to accelerate their convergence. Residual learning \cite{he2016deep} allows for much deeper networks to learn effective mappings without overfitting. Techniques such as pruning and weight compression  \cite{han2015deep} are becoming more commonplace. As problems become even larger and more complex, these techniques are increasingly being relied upon for efficient training and inference.

We approach this problem at the level of the image representation. JPEG is the most widespread image file format. Traditionally, the first step in using JPEGs for machine learning is to decompress them. We propose to skip this step and instead reformulate the ResNet architecture to perform its operations directly on compressed images. The goal is to produce a new network that is mathematically equivalent to the spatial domain network, but which operates on compressed images by including the compression transform into the network weights, which can be done because they are both linear maps. Because the ReLu function is non-linear, we develop an approximation technique for it. This is a general method and, to our knowledge, is the first attempt at formulating a piecewise linear function in the transform domain.

The contributions of this work are as follows
\begin{enumerate}
    \item The general method for expressing convolutional networks in the JPEG domain
    \item Concrete formulation for residual blocks to perform classification
    \item A model conversion algorithm to apply pretrained spatial domain networks to JPEG images
    \item Approximated Spatial Masking: the first general technique for application of piecewise linear functions in the transform domain
\end{enumerate}
By skipping the decompression step and by operating on the compressed format, we show a notable increase in speed for testing and a marginal speed for training.

\section{Prior Work}

We review prior work separated into three categories: compressed domain operations, machine learning in the compressed domain, and deep learning in the compressed domain.

\subsection{Compressed Domain Operations}

The expression of common operations in the compressed domain was an extremely active area of study in the late 80s and early 90s, motivated by the lack of computing power to quickly decompress, process, and recompress images and video. For JPEG, Smith and Rowe \cite{smith1993algorithms} formulate fast JPEG compatible algorithms for performing scalar and pixelwise addition and multiplication. This was extended by Shen and Sethi \cite{shen1995inner} to general blockwise operations and by Smith \cite{smith1994fast} to arbitrary linear maps. Natarajan and Vasudev \cite{natarajan1995fast} additionally formulate an extremely fast approximate algorithm for scaling JPEG images. For MPEG, Chang \etal \cite{chang1992video} introduce the basic algorithms for manipulating compressed video. Chang and Messerschmitt \cite{chang1993new} give a fast algorithm for decoding motion compensation before DCT which allows arbitrary video compositing operations to be performed. 

\subsection{Machine Learning in the Compressed Domain}

Compressed domain machine learning grew out of the work in the mid 90s. Arman \etal \cite{arman1993image} give the basic framework for image processing of compressed images. Feng and Jiang \cite{feng2002jpeg} show how image retrieval can be performed directly on compressed JPEGs. He \etal \cite{he2009efficient} extend their work with a hypothesis testing technique. Wu \etal \cite{wu2013sift} formulate the popular SIFT feature extraction in the DCT domain.

\subsection{Deep Learning in the Compressed Domain}

Because deep networks are non-linear maps, deep learning has received limited study in the compressed domain. Ghosh and Chellappa \cite{ghosh2016deep} use a DCT as part of their network's first layer and show that it speeds up convergence for training. This is extended by Ulicny \etal \cite{ulicny2018harmonic} to create separate filters for each DCT basis function. Wu \etal \cite{wu2018compressed} formulate a deep network for video action recognition that uses a separate network for i-frames and p-frames. Since the p-frame network functions on raw motion vectors and error residuals it is considered compressed domain processing, although it works in the spatial domain and not the quantized frequency domain as in this work. Wu \etal show a significant efficiency advantage compared to traditional 3D convolution architectures, which they attribute to the p-frame data being a minimal representation of the video motion. Gueguen \etal \cite{gueguen_2018_ICLR} formulate a traditional ResNet that operates on DCT coefficients directly instead of pixels, \eg the DCT coefficients are fed to the network. They show that learning converges faster on this input, further motivating the JPEG representation. 

\section{Background}

We briefly review the JPEG compression/decompression algorithm \cite{wallace1992jpeg} and introduce the multilinear method that we use to formulate our networks \cite{smith1994fast}.

\subsection{JPEG Compression}

The JPEG compression algorithm is defined as the following steps.

\begin{enumerate}
    \item Divide the image into $8 \times 8$ blocks
    \item Compute the 2D forward Discrete Cosine Transform (DCT Type 2) of each block
    \item Linearize the blocks using a zigzag order to produce a 64 component vector
    \item Element-wise divide each vector by a quantization coefficient
    \item Round the the vector elements to the nearest integer
    \item Run-length code and entropy code the vectors
\end{enumerate}

This process is repeated independently for each image plane. In most cases, the original image is transformed from the RGB color space to YUV and chroma subsampling is applied since the human visual system is less sensitive to small color changes than to small brightness changes \cite{winkler2001vision}. The decompression algorithm is the inverse process. Note that the rounding step (step 5) must be skipped during decompression. This is the step in JPEG compression where information is lost and is the cause of artifacts in decompressed JPEG images.

The magnitude of the information loss can be tuned using the quantization coefficients. If a larger coefficient is applied in step 4, then the result will be closer to 0 which increases its likelihood of being dropped altogether during rounding. In this way, the JPEG transform forces sparsity on the representation, which is why it compresses image data so well. This is coupled with the tendency of the DCT to push the magnitude of the coefficients into the upper left corner (the DC coefficient and the lowest spatial frequency) resulting in high spatial frequencies being dropped. Not only do these high spatial frequencies contribute less response to the human visual system, but they are also the optimal set to drop for a least squares reconstruction of the original image:

\begin{theorem}[DCT Least Squares Approximation Theorem]
    Given a set of $N$ samples of a signal $X = \{x_0, ... x_N\}$, let $Y = \{y_0, ... y_N\}$ be the DCT coefficients of $X$. Then, for any $1 \leq m \leq N$, the approximation
    
    \begin{equation}
    p_m(t) = \frac{1}{\sqrt{n}}y_o + \sqrt{\frac{2}{n}}\sum_{k=1}^{m} y_k\cos\left(\frac{k(2t + 1)\pi}{2n}\right)
    \end{equation}
    of $X$ minimizes the least squared error
    \begin{equation}
    e_m = \sum_{i=0}^{n} (p_m(i) - x_i)^2
    \end{equation}
    \label{thm:dctls}
\end{theorem}

Theorem \ref{thm:dctls} states that a reconstruction using the $m$ lowest spatial frequencies is optimal with respect to any other set of $m$ spatial frequencies. Proof of Theorem \ref{thm:dctls} is given in the supplementary material.

\subsection{JPEG Linear Map}

\label{sec:backjlm}

A key observation of the JPEG algorithm, and the foundation of most compressed domain processing methods \cite{chang1992video, chang1993new, natarajan1995fast, shen1995inner, shen1996direct, shen1998block, smith1993algorithms, smith1994fast} is that steps 1-4 of the JPEG compression algorithm are linear maps, so they can be composed, along with other linear operations, into a single linear map which performs the operations on the compressed representation. Step 5, the rounding step, is irreversible and ignored by decompression. Step 6, the entropy coding, is a nonlinear map and its form is computed from the data directly, so it is difficult to work with this representation. We define the JPEG Transform Domain as the output of Step 4 in the JPEG encoding algorithm. This is a standard convention of compressed domain processing. Inputs to the algorithms described here will be JPEGs after reversing the entropy coding.

Formally, we model a single plane image as the type (0, 2) tensor $I \in H^* \otimes W^*$ where $H$ and $W$ are vector spaces and $*$ denotes the dual space. We always use the standard orthonormal basis for these vector spaces which allows the free raising and lowering of indices without the use of a metric tensor.
We define the JPEG transform as the type (2, 3) tensor $J \in H \otimes W \otimes X^* \otimes Y^* \otimes K^*$. $J$ represents a linear map $J: H^* \otimes W^* \rightarrow X^* \otimes Y^* \otimes K^*$ and is computed as (in Einstein notation) 

\begin{equation}
I'_{xyk} = J^{hw}_{xyk}I_{hw}
\end{equation}
We say that $I'$ is the representation of $I$ in the JPEG transform domain. The indices $h,w$ give pixel position, $x,y$ give block position, and $k$ gives the offset into the block.

The form of $J$ is constructed from the JPEG compression steps listed in the previous section. Let the linear map $B: H^* \otimes W^* \rightarrow X^* \otimes Y^* \otimes M^* \otimes N^*$ be defined as 

\begin{equation}
B^{hw}_{xymn} = \left\{ \begin{array}{lr} 1 & \text{$h,w$ belongs in block $x,y$ at offset $m,n$} \\ 0 & \text{otherwise} \end{array} \right.
\end{equation} 
then $B$ can be used to break the image represented by $I$ into blocks of a given size such that the first two indices $x,y$ index the block position and the last two indices $m,n$ index the offset into the block.

Next, let the linear map $D: M^* \otimes N^* \rightarrow A^* \otimes B^*$ be defined as 

\begin{align}
D^{mn}_{\alpha\beta} = \frac{1}{4}V(\alpha)V(\beta)\cos\left(\frac{(2m+1)\alpha\pi}{16}\right)\cos\left(\frac{(2n+1)\beta\pi}{16}\right)
\end{align}
where $V(u)$ is a normalizing scale factor. Then $D$ represents the 2D discrete forward (and inverse) DCT. Let $Z: A^* \otimes B^* \rightarrow \Gamma^*$ be defined as 

\begin{equation}
Z^{\alpha\beta}_\gamma = \left\{ \begin{array}{lr} 1 & \text{$\alpha, \beta$ is at $\gamma$ under zigzag ordering} \\ 0 & \text{otherwise} \end{array} \right.
\end{equation}
then $Z$ creates the zigzag ordered vectors. Finally, let $S: \Gamma^* \rightarrow K^*$ be

\begin{equation}
    S^\gamma_k = \frac{1}{q_k}
\end{equation}
where $q_k$ is a quantization coefficient. This scales the vector entries by the quantization coefficients.

With linear maps for each step of the JPEG transform, we can then create the $J$ tensor described at the beginning of this section

\begin{equation}
J^{hw}_{xyk} = B^{hw}_{xymn}D^{mn}_{\alpha\beta}Z^{\alpha\beta}_{\gamma}S^\gamma_k
\end{equation}

The inverse mapping also exists as a tensor $\widetilde{J}$ which can be defined using the same linear maps with the exception of $S$. Let $\widetilde{S}$ be

\begin{equation}
\widetilde{S}^k_\gamma = q_k
\end{equation}
Then 

\begin{equation}
    \widetilde{J}^{xyk}_{hw} = B_{hw}^{xymn}D_{mn}^{\alpha\beta}Z_{\alpha\beta}^{\gamma}\widetilde{S}^k_\gamma
\end{equation}

Next consider a linear map $C: H^* \otimes W^* \rightarrow H^* \otimes W^*$ which performs an arbitrary pixel manipulation on an image plane $I$. To apply this mapping to a JPEG image $I'$, we first decompress the image, apply $C$ to the result, then compress that result to get the final JPEG. Since compressing is an application of $J$ and decompressing is an application of $\widetilde{J}$, we can form a new linear map $\Xi: X^* \otimes Y^* \otimes K^* \rightarrow X^* \otimes Y^* \otimes K^*$ as

\begin{equation}
\label{eq:stoj}
    \Xi^{xyk}_{x'y'k'} = \widetilde{J}^{xyk}_{hw}C^{hw}_{h'w'}J^{h'w'}_{x'y'k'}
\end{equation}
which applies $C$ in the JPEG transform domain. There are two important points to note about $\Xi$. The first is that, although it encapsulates decompression, applying $C$ and compressing, it uses far fewer operations than doing these processes separately since the coefficients are multiplied out. The second is that it is mathematically equivalent to performing $C$ on the decompressed image and compressing the result. It is not an approximation.

\section{JPEG Domain Residual Networks}

The ResNet architecture, consists of blocks of four basic operations: Convolution (potentially strided), ReLu, Batch Normalization, and Component-wise addition, with the blocks terminating with a global average pooling operation \cite{he2016deep} before a fully connected layer performs the final classification. Our goal will be to develop JPEG domain equivalents to these five operations. Network activations are given as a single tensor holding a batch of multi-channel images, that is $I \in N^* \otimes P^* \otimes H^* \otimes W^*$.

\subsection{Convolution}

The convolution operation follows directly from the discussion in Section \ref{sec:backjlm}. The convolution operation is a shorthand notation for a linear map $C: N^* \otimes P^* \otimes H^* \otimes W^* \rightarrow N^* \otimes P^* \otimes H^* \otimes W^*$. Since the same operation is applied to each image in the batch, we can represent $C$ with a type (3, 3) tensor. The entries of this tensor give the coefficient for a given pixel in a given input channel for each pixel in each output channel. We now develop the algorithm for representing discrete convolutional filters using this data structure.

A naive algorithm can simply copy randomly initialized convolution weights into this larger structure following the formula for convolution and then apply the JPEG compression and decompression tensors to the result. However, this is difficult to parallelize and incurs additional memory overhead to store the spatial domain operation. A more efficient algorithm would produce the JPEG domain operation directly and be easy to express as a compute kernel for a GPU. Start by considering the JPEG decompression tensor $\widetilde{J}$. Note that since $\widetilde{J} \in X \otimes Y \otimes K \otimes H^* \otimes W^*$ the last two indices of $\widetilde{J}$ form single channel image under our image model (\eg the last two indices are in $H^* \otimes W^*$). If the convolution can be applied to this "image", then the resulting map would decompress and convolve simultaneously. We can formulate a new tensor $\widehat{J} \in N \otimes H^* \otimes W^*$
by reshaping $\widetilde{J}$ and treating this as a batch of images \footnote{Consider as a concrete example using $32 \times 32$ images. Then $\widetilde{J}$ is of shape $4 \times 4 \times 64 \times 32 \times 32$ and the described reshaping gives $\widehat{J}$ of shape $1024 \times 1 \times 32 \times 32$ which can be treated as a batch of size 1024 of $32 \times 32$ images for convolution.}. Then, given randomly initialized filter weights, $K$ computing 
\begin{equation}
\label{eq:sdtojdc}
\widehat{C}^b = K \star \widehat{J}^b 
\end{equation}
where $\star$ indicates the convolution operation and $\widehat{J}^b$ indexes $\widehat{J}$ in the batch dimension, gives us the desired map. After reshaping $\widehat{C}$ back to the original shape of $\widetilde{J}$ to give $\widetilde{C}$, the full compressed domain operation can be expressed as 
\begin{equation}
\Xi^{pxyk}_{p'x'y'k'} = \widetilde{C}^{pxyk}_{p'hw}J^{hw}_{x'y'k'}
\end{equation}
where $p$ and $p'$ index the input and output channels of the image respectively. This algorithm skips the overhead of computing the spatial domain map explicitly and depends only on the batch convolution operation which is available in all GPU accelerated deep learning libraries. Further, the map can be precomputed to speed up inference by avoiding repeated applications of the convolution. At training time, the gradient of the compression and decompression operators is computed and used to find the gradient of the original convolution filter with respect to the previous layers error, then the map $\Xi$ is updated using the new filter. So, while inference efficiency of the convolution operation is greatly improved, training efficiency is limited by the more complex update. We show in Section \ref{sec:expeff} that the training throughput is still higher than the equivalent spatial domain model.

\subsection{ReLu}

\begin{figure}
    \centering
    \begin{subfigure}{0.20\textwidth}
        \captionsetup{width=.8\linewidth}
        \centering
        \includegraphics[width=\textwidth]{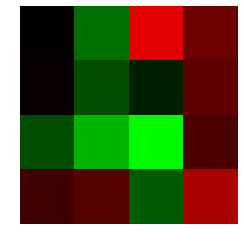}
        \caption{Original image}
    \end{subfigure}%
    \begin{subfigure}{0.2\textwidth}
        \captionsetup{width=.8\linewidth}
        \centering
        \includegraphics[width=\textwidth]{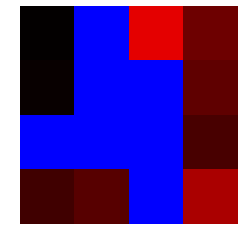}
        \caption{True ReLu}
    \end{subfigure}\\
    \begin{subfigure}{0.2\textwidth}
        \captionsetup{width=.8\linewidth}
        \centering
        \includegraphics[width=\textwidth]{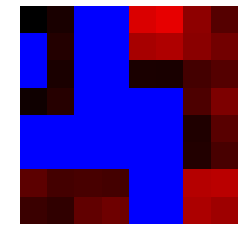}
        \caption{ReLu using direct approximation}
    \end{subfigure}%
    \begin{subfigure}{0.2\textwidth}
        \captionsetup{width=.8\linewidth}
        \centering
        \includegraphics[width=\textwidth]{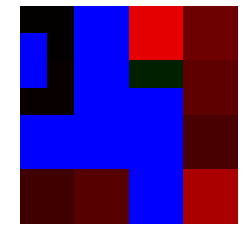}
        \caption{ReLu using ASM approximation}
    \end{subfigure}
    \caption{Example of ASM ReLu on an $8 \times 8$ block. Green pixels are negative, red pixels are positive, and blue pixels are zero. 6 spatial frequencies are used for both approximations. Note that the direct approximation fails to preserve positive pixel values.}
    \label{fig:asm}
\end{figure}

Computing ReLu in the JPEG domain is not as straightforward since ReLu is a non-linear function. Recall that the ReLu function is given by 

\begin{equation}
    r(x) = \begin{cases}
    x & x > 0 \\
    0 & x \leq 0
    \end{cases}
\end{equation}
We begin by defining the ReLu in the DCT domain and show how it can be trivially extended to the JPEG transform domain. To do this, we develop a general approximation technique called Approximated Spatial Masking that can apply any piecewise linear function to JPEG compressed images.

To develop this technique we must balance two seemingly competing criteria. The first is that we want to use the JPEG transform domain, since it has a computational advantage over the spatial domain. The second is that we want to compute a non-linear function which is incompatible with the JPEG transform. Can we balance these two constraints by sacrificing a third criterion? Consider an approximation of the spatial domain image that uses only a subset of the DCT coefficients. Computing this is fast, since it does not use the full set of coefficients, and gives us a spatial domain representation which is compatible with the non-linearity. What we sacrifice is accuracy. The accuracy-speed tradeoff is tunable to the problem by changing the size of the set of coefficients.

By Theorem \ref{thm:dctls} we use the lowest $m$ frequencies for an optimal reconstruction. For the $8 \times 8$ DCT used in the JPEG algorithm, this gives 15 spatial frequencies total (numbered 0 to 14). We can then fix a maximum number of spatial frequencies $k$ and use all coefficients $\phi$ such that $\phi \leq k$ as our approximation. 

If we now compute the piecewise linear function on this approximation directly there are two major problems. The first is that, although the form of the approximation is motivated by a least squares minimization, it is by no means guaranteed to reproduce the original values of \textit{any} of the pixels. The second is that this gives the value of the function in the spatial domain, and to continue using a JPEG domain network we would need to compress the result which adds computational overhead.

To solve the first problem we examine the intervals that the linear pieces fall into. The larger these intervals are, the more likely we are to have produced a value in the correct interval \footnote{For example if the original pixel value was 0.7 and the approximate value is 0.5, then the approximation is in the correct interval for ReLU ($\geq 0$) but its value is incorrect.} in our approximation. Further, since the lowest $k$ frequencies minimize the least squared error, the higher the frequency, the less likely it is to push a pixel value out of the correct range. With this motivation, we can produce a binary mask for each piece of the function. The linear pieces can then be applied directly to the DCT coefficients, and then multiplied by their masks and summed to give the final result. This preserves all pixel values. The only errors would be in the mask which would result in the wrong linear piece being applied. This is the fundamental idea behind the Approximated Spatial Masking (ASM) technique.

The final problem is that we now have a mask in the spatial domain, but the original image is in the DCT domain. There is a well known algorithm for pixelwise multiplication of two DCT images \cite{smith1993algorithms}, but it would require the mask to also be in the DCT domain. Fortunately, there is a straightforward solution that is a result of the multilinear analysis given in Section \ref{sec:backjlm}. Consider the bilinear map 
\begin{equation}
H: A^* \otimes B^* \times M^* \otimes N^* \rightarrow A^* \otimes B^*
\end{equation}
that takes a DCT block, $F$, and a spatial mask $G$, and produces the masked DCT block by pixelwise multiplication. Our task will be to derive the form of $H$. We proceed by construction. The steps of such an algorithm naively would be 
\begin{enumerate}
    \item Take the inverse DCT of F: $I_{mn} = D^{\alpha\beta}_{mn}F_{\alpha\beta}$
    \item Pixelwise multiply: $I'_{mn} = I_{mn}G_{mn}$
    \item Take the DCT of $I'$: $F'_{\alpha'\beta'} = D^{mn}_{\alpha'\beta'}I'_{mn}$.
\end{enumerate}
Since these three steps are linear or bilinear maps, they can be combined 
\begin{equation}
F'_{\alpha'\beta'} = F^{\alpha\beta}[D^{mn}_{\alpha\beta}D^{mn}_{\alpha'\beta'}]G_{mn}
\end{equation}
Giving the final bilinear map $H$ as 
\begin{equation}
H^{\alpha\beta mn}_{\alpha'\beta'} = D^{\alpha\beta mn}D^{mn}_{\alpha'\beta'}
\end{equation}

We call $H$ the Harmonic Mixing Tensor since it gives all the spatial frequency permutations that we need. $H$ can be precomputed to speed up computation.

To use this technique to compute the ReLu function, consider this alternative formulation
\newcommand{\nnm}{\mathrm{nnm}}

\begin{equation}
    \nnm(x) = \begin{cases}
    1 & x > 0 \\
    0 & x \leq 0
    \end{cases}
\end{equation}
We call the function $\nnm(x)$ the nonnegative mask of $x$. This is our binary mask for ASM. We express the ReLu function as 

\begin{equation}
    r(x) = \nnm(x)x
\end{equation}
This new function can be computed efficiently from fewer spatial frequencies with much higher accuracy since only the sign of the original function needs to be correct. Figure \ref{fig:asm} gives an example of this algorithm on a random block and compares it to computing ReLu on the approximation directly. Note that in the ASM image the pixel values of all positive pixels are preserved, the only errors are in the mask. In the direct approximation, however, none of the pixel values are preserved and it suffers from masking errors. The magnitude of the error is tested in Section \ref{sec:exprla} and pseudocode for the ASM algorithm is given in the supplementary material. 

To extend this method from the DCT domain to the JPEG transform domain, the rest of the missing JPEG tensor can simply be applied as follows:
\begin{equation}
H^{k mn}_{k'} = Z^{k}_{\gamma} \widetilde{S}^{\gamma}_{\alpha\beta} D^{\alpha\beta mn}D^{mn}_{\alpha'\beta'} S^{\alpha'\beta'}_{\gamma'} Z^{\gamma'}_{k'}
\end{equation}
Since the operation is the same for each block, and there are no interactions between blocks, the blocking tensor $B$ can be skipped.

\subsection{Batch Normalization}
\label{sec:jdrbn}

Batch normalization \cite{ioffe2015batch} has a simple and efficient formulation in the JPEG domain. Recall that batch normalization defines two learnable parameters: $\gamma$ and $\beta$. A given feature map $I$ is first centered and then normalized over the batch, then scaled by $\gamma$ and translated by $\beta$. The full formula is
\begin{equation}
\text{BN}(I) = \gamma\frac{I - \e[I]}{\sqrt{\var[I]}} + \beta
\end{equation}
So to define the batch normalization operation in the JPEG domain, we need four parts: the mean, the variance, scalar multiplication and scalar addition. Again, we first derive the result in the DCT domain and trivially extend to the JPEG transform domain.

We start with the sample mean. Observe, from the definition of the DCT, the first DCT coefficient is given by 
\begin{equation}
D_{00} = \frac{1}{2\sqrt{2N}}\sum_{x=0}^N\sum_{y=0}^N I_{xy}
\end{equation}
In other words, the (0,0) DCT coefficient is proportional to the mean of the block. Further, since the DCT basis is orthonormal, we can be sure that the remaining DCT coefficients do not depend on the mean. This means that to center the image we need only set the (0,0) DCT coefficient to 0. For tracking the running mean, we simply read this value. Note that this is a much more efficient operation than the mean computation in the spatial domain.

Next, to get the variance, we use the following theorem:
\begin{theorem}[The DCT Mean-Variance Theorem]
    Given a set of samples of a signal $X$ such that $\e[X] = 0$, let $Y$ be the DCT coefficients of $X$. Then 
    \begin{equation}
    \var[X] = \e[Y^2]
    \end{equation}
\end{theorem}
Intuitively this makes sense because the (0,0) coefficient represents the mean, the remaining DCT coefficients are essentially spatial oscillations around the mean, which should define the variance. Proof of this theorem is given in the supplementary material.

To apply $\gamma$ and the variance, we use scalar multiplication. Since JPEG is linear, this is unchanged
\begin{equation}
    J(\gamma I) = \gamma J(I)
\end{equation}
For scalar addition to apply $\beta$, note that since the (0,0) coefficient is the mean, and adding $\beta$ to every pixel in the image is equivalent to raising the mean by $\beta$, we can simply add $\beta$ to the (0,0) coefficient.

To extend this to JPEG is simple. The proportionality constant for the (0,0) coefficient is $\frac{1}{2\sqrt{2 \times 8}} = \frac{1}{8}$. For this reason, many quantization matrices use $8$ as the (0,0) quantization coefficient. This means that the 0th block entry for a block does not need any proportionality constant, it stores exactly the mean. So for adding $\beta$, we can simply set the 0th position to $\beta$ without performing additional operations. The other operations are unaffected. 

\subsection{Component-wise Addition}

Component-wise addition is the simplest formulation in our network. This is a well known result detailed in \cite{chang1992video, shen1998block, shen1995inner, smith1993algorithms} among others. Since the JPEG transform, $J$, is a linear map, for two images $F$ and $G$, we have 
\begin{equation}
J(F + G) = J(F) + J(G)
\end{equation}
meaning that we can simply perform a component-wise addition of the JPEG compressed results with no need for further processing.

\subsection{Global Average Pooling}

\begin{figure}
    \centering
    \includegraphics[width=0.5\linewidth]{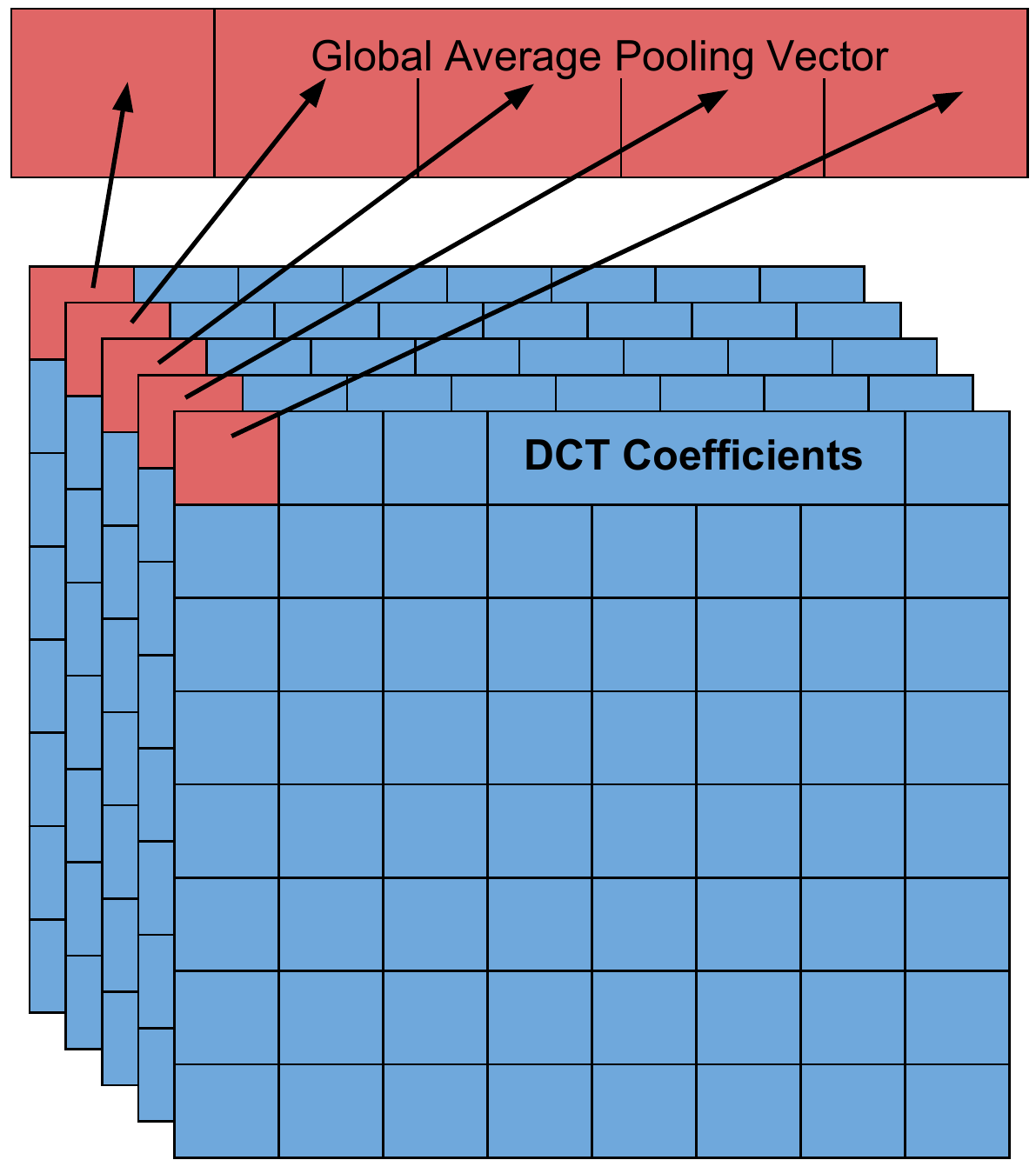}
    \caption{Global average pooling. The 0th coefficient of each block can be used directly with no computation.}
    \label{fig:gap}
\end{figure}

Global average pooling also has a simple formulation in JPEG domain. Recall from the discussion of Batch Normalization (Section \ref{sec:jdrbn}) that the 0th element of the block after quantization is equal to the mean of the block. Then this element can be extracted channel-wise from each block and the global average pooling result is the channel-wise mean of these elements.

Furthermore, our network architecture for classification will always reduce the input images to a single block, which can then have its mean extracted and reported as the global average pooling result directly. Note the efficiency of this process: rather than channel-wise averaging in a spatial domain network, we simply have an unconditional read operation, one per channel. This is illustrated in Figure \ref{fig:gap}.

\subsection{Model Conversion}

The previous sections described how to build the ResNet component operations in the JPEG domain. While this implies straightforward algorithms for both inference and learning on JPEGs, we can also convert pre-trained models for JPEG inference. This allows any model that was trained on spatial domain images to benefit from our algorithms at inference time. Consider Equation \ref{eq:sdtojdc}. In this equation, $K$ holds the randomly initialized convolution filter. By instead using pretrained spatial weights for $K$, the convoltion will work as expected on JPEGs. Similarly, pretrained $\alpha, \beta, \mu, \sigma$ for batch normalization can be provided. By doing this for each layer in a pretrained network, the network will operate on JPEGs. The only caveat is that the ReLu approximation accuracy can effect the final performance of the network since the weights were not trained to cope with it. This is tested in Section \ref{sec:exprla}.

\section{Experiments}

We give experimental evidence for the efficacy of our method, starting with a discussion of the architectures we use and the datasets. We use model conversion as a sanity check, ensuring that the JPEG model with exact ReLu matches exactly the testing accuracy of a spatial domain model. Next we show how the ReLu approximation accuracy effects overall network performance. We conclude by showing the training and testing time advantage of our method.

\subsection{Network Architectures and Datasets}

\begin{figure}
    \centering
    \includegraphics[width=\linewidth]{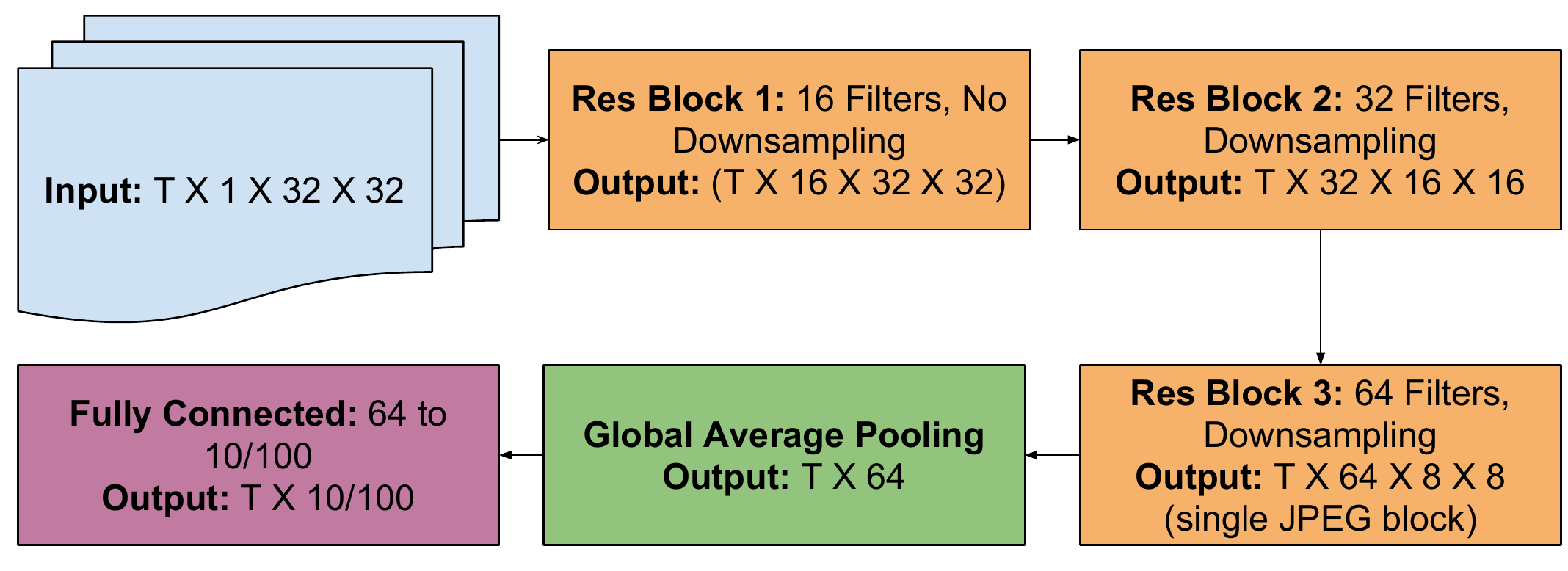}
    \caption{Simple network architecture. $T$ indicates the batch size.}
    \label{fig:na}
\end{figure}

Since we are concerned with reproducing the inference results of spatial domain networks, we choose the MNIST \cite{lecun1998mnist} and CIFAR-10/100 \cite{krizhevsky2009learning} datasets since they are easy to work with. The MNIST images are padded to $32 \times 32$ to ensure an even number of JPEG blocks. Our network architecture is shown in Figure \ref{fig:na}. The classification network consists of three residual blocks with the final two performing downsampling so that the final feature map consists of a single JPEG block. The goal of this architecture is not to get high accuracy, but rather to serve as a point of comparison for the spatial and JPEG algorithms.

\subsection{Model Conversion}

For this first experiment, we show empirically that the JPEG formulation is mathematically equivalent to the spatial domain network. To show this, we train 100 spatial domain models on each of the three datasets and give their mean testing accuracies. We then use model conversion to transform the pretrained models to the JPEG domain and give the mean testing accuracies of the JPEG models. The images are losslessly JPEG compressed for input to the JPEG networks and the exact (15 spatial frequency) ReLu formulation is used. The result of this test is given in Table \ref{tab:mc}. Since the accuracy difference between the networks is extremely small, the deviation is also included. 

\begin{table}[h]
    \centering
    \begin{tabular}{|r|l|l|l|}
        \hline
        Dataset & Spatial & JPEG & Deviation \\ \hline
        MNIST & 0.988 & 0.988 & 2.999e-06 \\ \hline
        CIFAR10 & 0.725 & 0.725 & 9e-06 \\ \hline
        CIFAR100 & 0.385 & 0.385 & 1e-06 \\ \hline
    \end{tabular}
    \caption{Model conversion accuracies. Spatial and JPEG testing accuracies are the same to within floating point error.}
    \label{tab:mc}
\end{table}

\subsection{ReLu Approximation Accuracy}
\label{sec:exprla}

\begin{figure*}
    \centering
    \caption{ReLu accuracy results.}
    \begin{subfigure}{0.33\textwidth}
        \captionsetup{width=.8\linewidth}
        \centering
        \includegraphics[width=\textwidth]{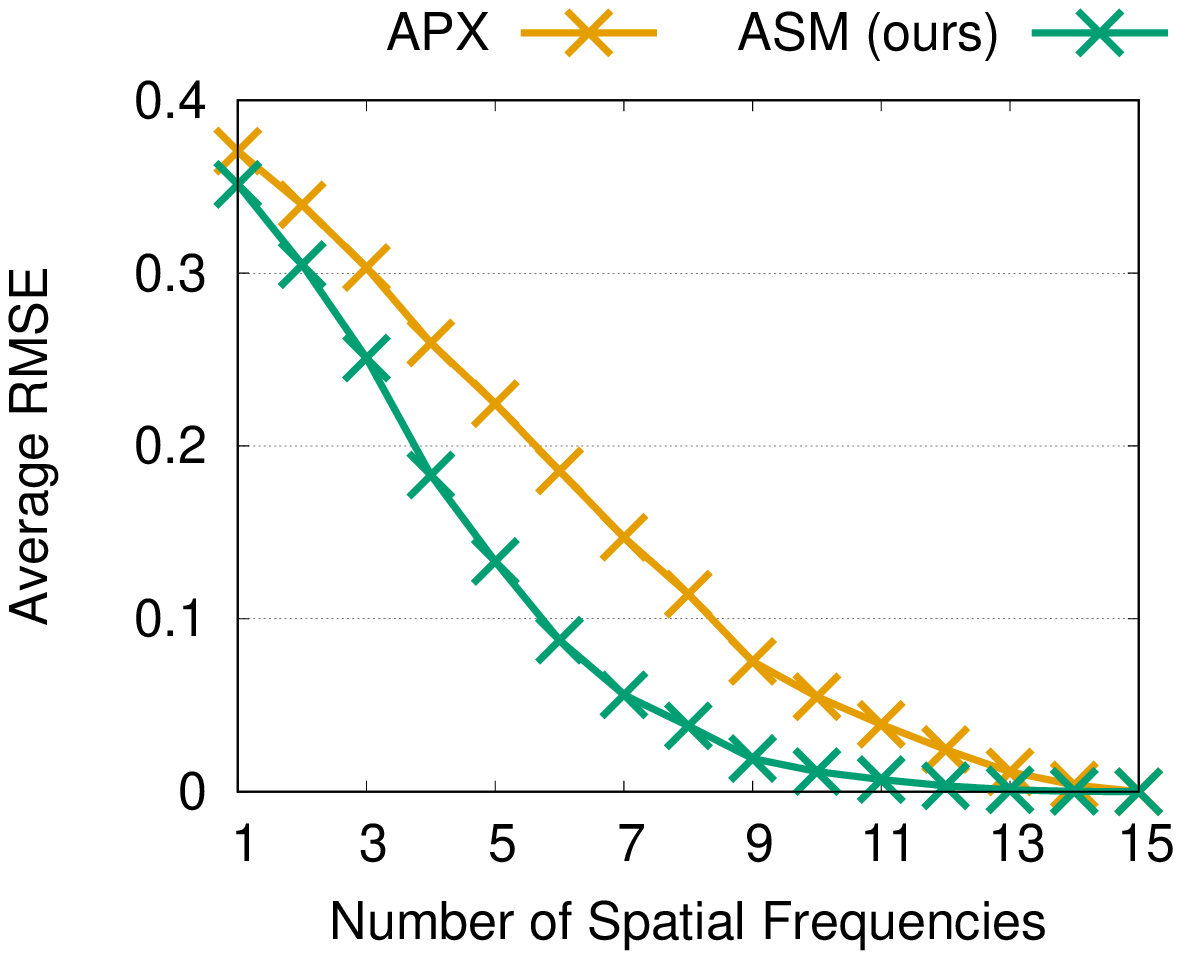}
        \caption{ReLu blocks error. Our ASM method consistently gives lower error than the naive approximation method. }
        \label{fig:rba}
    \end{subfigure}%
    \begin{subfigure}{0.33\textwidth}
        \captionsetup{width=.8\linewidth}
        \centering
        \includegraphics[width=\textwidth]{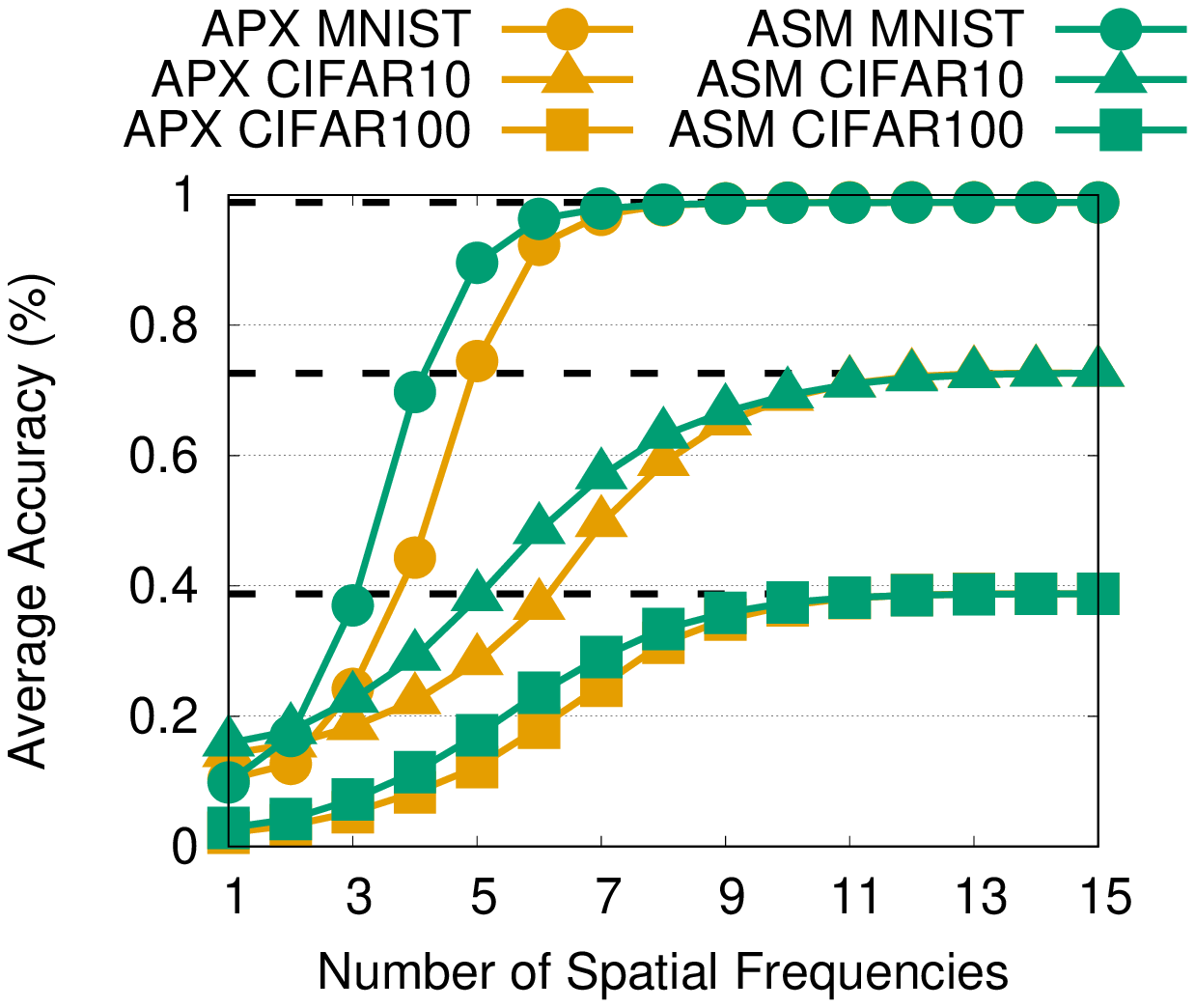}
        \caption{ReLu model conversion accuracy. ASM again outperforms the naive approximation. The spatial domain accuracy is given for each dataset with dashed lines.}
        \label{fig:ra}
    \end{subfigure}%
    \begin{subfigure}{0.33\textwidth}
        \captionsetup{width=.8\linewidth}
        \centering
        \includegraphics[width=\textwidth]{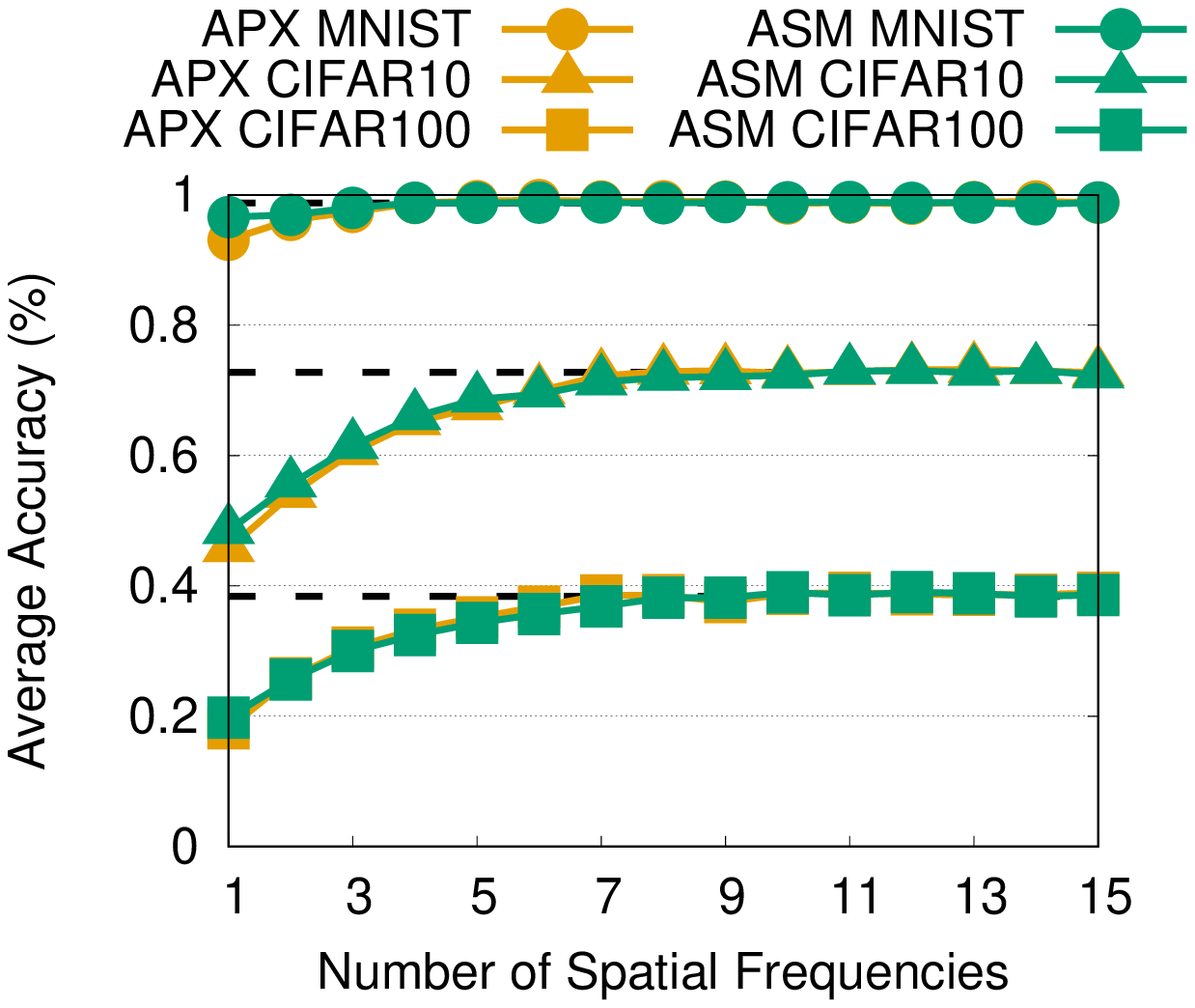}
        \caption{ReLu training accuracy. The network weights have learned to correct for the ReLu approximation allowing fewer spatial frequencies to be used for high accuracy.}
        \label{fig:rt}
    \end{subfigure}
\end{figure*}

Next, we examine the impact of the ReLu approximation. We start by examining the raw error on individual $8 \times 8$ blocks. For this test, we take random $4 \times 4$ pixel blocks in the range $[-1, 1]$ and scale them to $8 \times 8$ using a box filter. Fully random $8 \times 8$ blocks do not accurately represent the statistics of real images and are known to be a worst case for the DCT transform. The $4 \times 4$ blocks allow for a large random sample size while still approximating real image statistics. We take 10 million blocks and compute the average RMSE of our ASM technique and compare it to computing ReLu directly on the approximation (APX). This test is repeated for all one to fifteen spatial frequencies. The result, shown in Figure \ref{fig:rba} shows that our ASM method gives a better approximation (lower RMSE) through the range of spatial frequencies. 

This test provides a strong motivation for the ASM method, so we move on to testing it in the model conversion setting. For this test, we again train 100 spatial domain models and then perform model conversion with the ReLu layers ranging from 1-15 spatial frequencies. We again compare our ASM method with the APX method. The result is given in Figure \ref{fig:ra}. Again the ASM method outperforms the APX method.

As a final test, we show that if the models are trained in the JPEG domain, the CNN weights will actually learn to cope with the approximation and fewer spatial frequencies are required for good accuracy. The result in Figure \ref{fig:rt} shows that the ASM method again outperforms the APX method and that the network weights have learned to cope with the approximation.

\subsection{Efficiency of Training and Testing}
\label{sec:expeff}

\begin{figure}[b]
    \includegraphics[width=\linewidth]{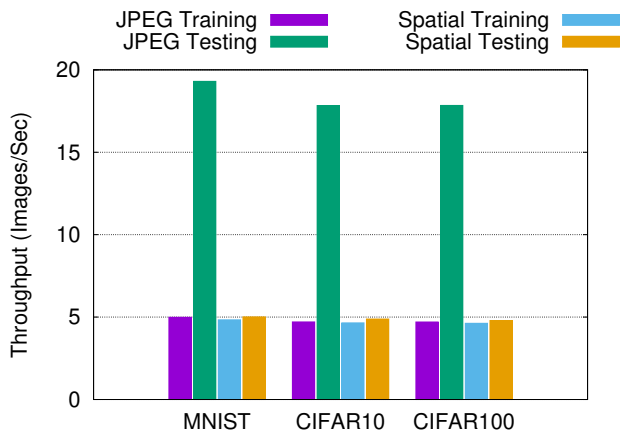}
    \caption{Throughput. The JPEG model has a more complex gradient which limits speed improvement during training. Inference, however,  sees considerably higher throughput.}
    \label{fig:tp}
\end{figure}    

Finally, we show the throughput for training and testing. For this we test on all three datasets by training and testing a spatial model and training and testing a JPEG model and measuring the time taken. This is then converted to an average throughput measurement. The experiment is performed on an NVIDIA Pascal GPU with a batch size of 40 images. The results, shown in Figure \ref{fig:tp}, show that the JPEG model is able to outperform the spatial model in all cases, but that the performance on training is still limited. This is caused by the more complex gradient created by the convolution and ReLu operations. At inference time, however, performance is greatly improved over the spatial model.

\section{Conclusion and Future Work}

In this work we showed how to formulate deep residual learning in the JPEG transform domain, and that it provides a notable performance benefit in terms of processing time per image. Our method expresses convolutions as linear maps \cite{smith1994fast} and introduces a novel approximation technique for ReLu. We showed that the approximation can achieve highly performant results with little impact on classification accuracy.

Future work should focus on two main points. The first is efficiency of representation. Our linear maps take up more space than spatial domain convolutions. This makes it hard to scale the networks to datasets with large image sizes. Secondly, library support in commodity deep learning libraries for some of the features required by this algorithm are lacking. As of this writing, true sparse tensor support is missing in all of PyTorch \cite{paszke2017automatic}, TensorFlow \cite{tensorflow2015-whitepaper}, and Caffe \cite{jia2014caffe}, with these tensors being represented as coordinate lists which are known to be highly non-performant. Additionally, the \texttt{einsum} function for evaluating multilinear expressions is not fully optimized in these libraries when compared to the speed of convolutions in libraries like CuDNN \cite{chetlur2014cudnn}, though we make use of the \texttt{opt\_einsum} \cite{G2018opt} tool to partially mitigate this.

\section{Acknowledgment}

This research was partially funded by Facebook AI Research. We especially thank Dr. Ser-Nam Lim and his team at Facebook for their continued support of our work.

\newpage

{\small
    \printbibliography
}
\cleardoublepage


\section*{Supplementary material}

# Supplementary Material

## 1. Proof of the DCT Least Squares Approximation Theorem

**Theorem 1** (DCT Least Squares Approximation Theorem). *Given a set of $N$ samples of a signal $X = \{x_0, ...x_N\}$, let $Y = \{y_0, ...y_N\}$ be the DCT coefficients of $X$. Then, for any $1 \leq m \leq N$, the approximation*

$$p_m(t) = \frac{1}{\sqrt{n}} y_o + \sqrt{\frac{2}{n}} \sum_{k=1}^{m} y_k \cos\left(\frac{k(2t+1)\pi}{2n}\right) \quad (1)$$

*of $X$ minimizes the least squared error*

$$e_m = \sum_{i=0}^{n} (p_m(i) - x_i)^2 \quad (2)$$

*Proof.* First consider that since Equation 1 represents the Discrete Cosine Transform, which is a Linear map, we can write rewrite it as

$$D_m^T y = x \quad (3)$$

where $D_m$ is formed from the first $m$ rows of the DCT matrix, $y$ is a row vector of the DCT coefficients, and $x$ is a row vector of the original samples.

To solve for the least squares solution, we use the the normal equations, that is we solve

$$D_m D_m^T y = D_m x \quad (4)$$

and since the DCT is an orthonormal transformation, the rows of $D_m$ are orthogonal, so $D_m D_m^T = I$. Therefore

$$y = D_m x \quad (5)$$

Since there is no contradiction, the least squares solution must use the first $m$ DCT coefficients. □

## 2. Proof of the DCT Mean-Variance Theorem

**Theorem 2** (DCT Mean-Variance Theorem). *Given a set of samples of a signal $X$ such that $\mathrm{E}[X] = 0$, let $Y$ be the DCT coefficients of $X$. Then*

$$\mathrm{Var}[X] = \mathrm{E}[Y^2] \quad (6)$$

*Proof.* Start by considering $\mathrm{Var}[X]$. We can rewrite this as

$$\mathrm{Var}[X] = \mathrm{E}[X^2] - \mathrm{E}[X]^2 \quad (7)$$

Since we are given $\mathrm{E}[X] = 0$, this simplifies to

$$\mathrm{Var}[X] = \mathrm{E}[X^2] \quad (8)$$

Next, we express the DCT as a linear map such that $X = DY$ and rewrite the previous equation as

$$\mathrm{Var}[X] = \mathrm{E}[(DY)^2] \quad (9)$$

Squaring gives

$$\mathrm{E}[(DY)^2] = \mathrm{E}[(D^T D)Y^2] \quad (10)$$

Since $D$ is orthogonal this simplifies to

$$\mathrm{E}[(D^T D)Y^2] = \mathrm{E}[(D^{-1} D)Y^2] = \mathrm{E}[Y^2] \quad (11)$$

□

## 3. Algorithms

We conclude by outlining in pseudocode the algorithms for the three layer operations described in the paper. Algorithm 1 gives the code for convolution explosion, Algorithm 2 gives the code for the ASM ReLu approximation, and Algorithm 3 gives the code for Batch Normalization.

---

**Algorithm 1** Convolution Explosion. $K$ is an initial filter, $p, p'$ are the input and output channels, $h, w$ are the image height and width, $s$ is the stride, $\star_s$ denotes the discrete convolution with stride $s$. $J$ and $\widetilde{J}$ are constants of shape $(x, y, k, h, w)$ with $y = h/8$, $x = w/8$, $k = 64$.

---

   **function** EXPLODE($K, p, p', h, w, s$)
      $d_j \leftarrow \mathbf{shape}(\widetilde{J})$
      $d_b \leftarrow (d_j[0], d_j[1], d_j[2], 1, h, w)$
      $\widehat{J} \leftarrow \mathbf{reshape}(\widetilde{J}, d_b)$
      $\widehat{C} \leftarrow \widehat{J} \star_s K$
      $d_c \leftarrow (p, p', d_j[0], d_j[1], d_j[2], h/s, h/s)$
      $\widetilde{C} \leftarrow \mathbf{reshape}(\widehat{C}, d_c)$
      **return** $\widetilde{C}_{p'hw}^{pxyk} J_{x'y'k'}^{hw}$

**Algorithm 2** Approximated Spatial Masking for ReLu. $F$ is a DCT domain block, $\phi$ is the desired maximum spatial frequencies, $N$ is the block size.

---

**function** RELU($F, \phi, N$)
    $M \leftarrow \text{ANNM}(F, \phi, N)$
    **return** APPLYMASK($F, M$)
**function** ANNM($F, \phi, N$)
    $I \leftarrow \textbf{zeros}(N, N)$
    **for** $i \in [0, N)$ **do**
        **for** $j \in [0, N)$ **do**
            **for** $\alpha \in [0, N)$ **do**
                **for** $\beta \in [0, N)$ **do**
                      **if** $\alpha + \beta \leq \phi$ **then**
                          $I_{ij} \leftarrow I_{ij} + F_{ij} D_{ij}^{\alpha\beta}$
    $M \leftarrow \textbf{zeros}(N, N)$
    $M[I > 0] \leftarrow 1$
    **return** $M$
**function** APPLYMASK($F, M$)
    **return** $H_{\alpha'\beta'}^{\alpha\beta ij} F_{\alpha\beta} M_{ij}$

---

**Algorithm 3** Batch Normalization. $F$ is a batch of JPEG blocks (dimensions $N \times 64$), $S$ is the inverse quantization matrix, $m$ is the momentum for updating running statistics, $t$ is a flag that denotes training or testing mode. The parameters $\gamma$ and $\beta$ are stored externally to the function. $\widehat{\phantom{x}}$ is used to denote a batch statistic and $\widetilde{\phantom{x}}$ is used to denote a running statistic.

---

**function** BATCHNORM($F, S, m, t$)
    **if** $t$ **then**
        $\mu \leftarrow \textbf{mean}(F[:, 0])$
        $\widehat{\mu} \leftarrow F[:, 0]$
        $F[:, 0] = 0$
        $D_g \leftarrow F_k S_k$
        $\widehat{\sigma^2} \leftarrow \textbf{mean}(F^2, 1)$
        $\sigma^2 \leftarrow \textbf{mean}(\widehat{\sigma^2} + \widehat{\mu}^2) - \mu^2$
        $\widetilde{\mu} \leftarrow \widetilde{\mu}(1 - m) + \mu m$
        $\widetilde{\sigma^2} \leftarrow \widetilde{\sigma^2}(1 - m) + \mu m$
        $F[:, 0] \leftarrow F[:, 0] - \mu$
        $F \leftarrow \frac{\gamma F}{\sigma}$
        $F[:, 0] \leftarrow F[:, 0] + \beta$
    **else**
        $F[:, 0] \leftarrow F[:, 0] - \widetilde{\mu}$
        $F \leftarrow \frac{\gamma F}{\widetilde{\sigma}}$
        $F[:, 0] \leftarrow F[:, 0] + \beta$
    **return** $F$



\end{document}